%% file: main.tex
\documentclass[11pt,a4paper]{article}
\usepackage[]{emnlp2023}
\usepackage{times,latexsym}
\usepackage[T1]{fontenc}
\usepackage{microtype}

\usepackage{amsmath}
\usepackage{amsthm}
\usepackage{amsfonts}
\usepackage{amssymb}
\usepackage{mathtools}
\usepackage{tikz}
\usepackage{mathabx}

\usepackage{tikz}
\usetikzlibrary{bayesnet}
\usetikzlibrary{arrows}
\usetikzlibrary{decorations}
\usepackage{subcaption}

\newenvironment{itemizesquish}[2]{\begin{list}{\labelitemi}{\setlength{\itemsep}{#1}\setlength{\labelwidth}{#2}\setlength{\leftmargin}{\labelwidth}\addtolength{\leftmargin}{\labelsep}}}{\end{list}}

\definecolor{ForestGreen}{RGB}{34,139,34}
\usepackage{graphicx}
\usepackage{booktabs}
\usepackage{tabularx}
\usepackage{tabulary}
\usepackage{colortbl}
\usepackage{xspace}
\newcolumntype{Y}{>{\centering\arraybackslash}X}
  \newcolumntype{P}{>{\raggedleft\arraybackslash}X}
\usepackage{multirow}
\usepackage{stfloats}
\usepackage{soul}

\usepackage{algorithm}
\usepackage[noend]{algorithmic}

\usepackage[noabbrev,capitalise]{cleveref}

\DeclareRobustCommand{\hlgreen}[1]{{\sethlcolor{YellowGreen}\hl{#1}}}
\DeclareRobustCommand{\hlred}[1]{{\sethlcolor{Orange}\hl{#1}}}
\DeclareRobustCommand{\hlblue}[1]{{\sethlcolor{Cyan}\hl{#1}}}

\input{macros}

\title{Are Large Language Models Temporally Grounded?}

\author{$^1$Yifu Qiu~~~~$^1$Zheng Zhao~~~~$^1$Yftah Ziser~~~~\\$^2$\textbf{Anna Korhonen}~~~~$^1$\textbf{Edoardo M. Ponti}~~~~$^1$\textbf{Shay B. Cohen}\\
  $^1$Institute for Language, Cognition and Computation, University of Edinburgh \\
  $^2$Language Technology Lab, University of Cambridge \\
  \texttt{\{yifu.qiu,zheng.zhao,yftah.ziser,eponti,scohen\}@ed.ac.uk} \\
  }

\date{}

\begin{document}
\maketitle

\begin{abstract}
Are Large language models (LLMs) temporally grounded?
Since LLMs cannot perceive and interact with the environment, it is impossible to answer this question directly.
Instead, we provide LLMs with textual narratives and probe them with respect to their common-sense knowledge of the structure and duration of events, their ability to order events along a timeline, and self-consistency within their temporal model (e.g., temporal relations such as \textit{after} and \textit{before} are mutually exclusive for any pair of events). 
We evaluate state-of-the-art LLMs (such as LLaMA 2 and GPT-4) on three tasks reflecting these abilities. Generally, we find that LLMs lag significantly behind both human performance as well as small-scale, specialised LMs. In-context learning, instruction tuning, and chain-of-thought prompting reduce this gap only to a limited degree. Crucially, LLMs struggle the most with self-consistency, displaying incoherent behaviour in at least 27.23\% of their predictions. Contrary to expectations, we also find that scaling the model size does not guarantee positive gains in performance. To explain these results, we study the sources from which LLMs may gather temporal information: we find that sentence ordering in unlabelled texts, available during pre-training, is only weakly correlated with event ordering. Moreover, public instruction tuning mixtures contain few temporal tasks. Hence, we conclude that current LLMs lack a consistent temporal model of textual narratives.\footnote{ Code, datasets, and LLM outputs are available at \url{https://github.com/yfqiu-nlp/temporal-llms}.}

\end{abstract}

\section{Introduction}

\begin{figure*}
    \centering
    \includegraphics[width=\linewidth]{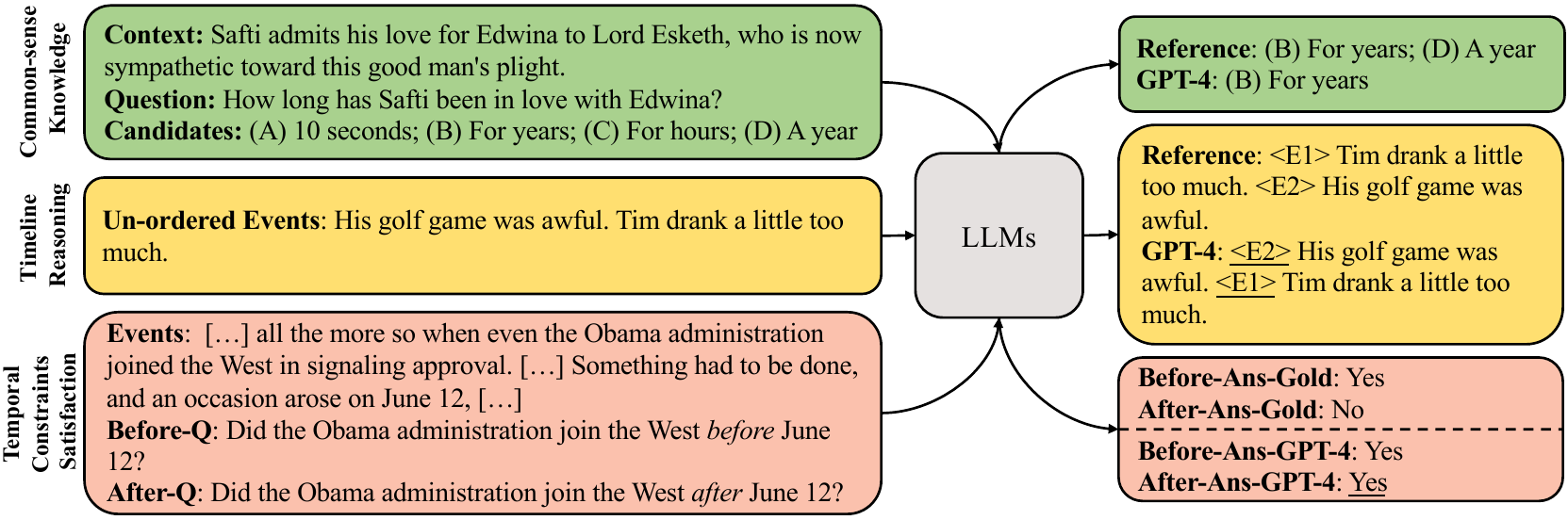}
    \caption{Examples from three datasets to evaluate the temporal grounding of LLMs based on common-sense knowledge about events, timeline reasoning, and temporal constraints satisfaction. We include the predictions from the best-performing model, GPT-4, with in-context learning. We highlight wrong predictions with \underline{underline}.}
    \label{fig:temporal-grounding-qualitative-examples}
\end{figure*}

Recent large language models (LLMs) like GPT-4 have shown impressive progress on various downstream tasks in natural language processing \cite{NEURIPS2020_1457c0d6-gpt3,chung2022scaling-flan-t5,workshop2023bloom,chowdhery2022palm,openai2023gpt4}.
However, a notorious deficiency of these models is the \textit{lack of grounding} \cite{yun-etal-2021-vision-language-model-grounding,abdou-etal-2021-language-model-color-grounding,carta2023grounding,mahowald2023dissociating}. LLMs, which are solely trained with textual data, acquire knowledge from distributional patterns of language. Humans, on the other hand, perceive and interact with the world through the dimensions of time and space.
Accurately grounding intelligent systems within the physical world is crucial for developing a versatile form of intelligence that is truly embodied \cite{dasguptacollaborating-embodied-reasoning,driess2023palme,carta2023grounding}, as well as capable of sample-efficient learning and causal discovery through mental simulations \citep{lecun2022path}.

In this work, we focus on temporal reasoning as a case study for establishing to what extent LLMs are grounded in the absence of perception and action. We first present a framework (\cref{sec:temporal-grounding-formalism}) for evaluating temporal grounding, where we decompose this property into three fundamental aspects: 1) commonsense knowledge about the typical duration, occurrence, frequency, and structure of events, 2) ability to order events along a timeline and 3) consistency based on temporal constraints. Then, we carefully choose and curate three benchmarks (\cref{sec:temporal-grounding-benchmarks}) for capturing each of these abilities: McTACO \cite{zhou-etal-2019-mctaco-dataset}, CaTeRS \cite{mostafazadeh-etal-2016-caters-dataset} and TempEvalQA \cite{llorens-etal-2015-TempEvalQA-dataset}, respectively. 

We conduct an extensive empirical evaluation of the temporal reasoning capabilities of state-of-the-art LLMs, including both open-source (e.g., LLaMA 2) and proprietary (e.g., GPT-4). In addition, we investigate the effect of recent techniques that facilitate generalisation to new tasks with ablations: we compare zero-shot inference, few-shot prompting using in-context learning, scaling in terms of model parameters and the number of few-shot examples, and chain-of-thought prompting.

Firstly, we find that all the tasks requiring temporal grounding remain extremely challenging for LLMs (\cref{sec:results}), even for the recently released state-of-the-art GPT-4. In particular, LLMs lag significantly behind human performance and even behind specialised, small-scale, fine-tuned language models in temporal commonsense knowledge and event ordering. Most strikingly, however, the results reported on our TempEvalQA-Bi benchmark (\cref{sec:tempeval-bi-results}) reveal that the predictions of LLMs are not consistent with respect to mutually exclusive temporal relations. For example, LLaMA-65B and \texttt{text-davinci-003} display a percentage of inconsistent predictions at 78.13\% and 69.64\%, respectively. Secondly, we observe that recent techniques introduced for boosting LLMs' performance, including few-shot in-context learning (\cref{sec:scaling-effects}) and chain of thought (\cref{sec:chain-of-thought}), have a positive but limited effect. Thirdly, scaling the examples for in-context learning and model size does not necessarily improve the model performance on temporal tasks (\cref{sec:scaling-effects}).

Finally, to understand the reason for the underwhelming performance of LLMs in temporal tasks, 
we attempt to determine whether enough temporal information is provided to them during their training pipeline (Section~\ref{sec:source-of-temporal-info}).
We find that in the unlabelled texts available during pre-training, sentence ordering and event ordering are weakly correlated. 
Hence, LLMs assign only a slightly higher probability to sentences describing a ground-truth event series in temporal order, rather than an inverted order. 
This is especially the case whenever explicit temporal markers (e.g., \textit{before} and \textit{after}) are missing. We also find that most training mixtures for instruction tuning lack temporal tasks.
Thus, temporal grounding cannot emerge from the current LLM training paradigm.

\section{Motivation and Framework} 
\label{sec:temporal-grounding-formalism}

While the question of whether language models contain a spatial map of the world has been thoroughly investigated \citep[\textit{inter alia}]{louwerse2009language,mikolov2013distributed,rahimi-etal-2017-neural,faisal2022geographic}, \textit{temporal} grounding of language models remains an under-explored area. Recently, \citet{gurnee2023language} found that a linear probe can recover the year of existence of an entity from its representation. From this evidence, they conclude that LLMs can learn (temporal) world models; however, this seems unwarranted given that `world model' defines the ability of grounded agents to internally simulate physical dynamics \citep{ha2018worldmodels}. On the other hand, since LLMs are incapable of action and perception, this ability cannot be assessed directly.

Instead, to study whether LLMs are temporally grounded, we we assess their ability to reason about textual narratives, which contain accounts of series of events. Events are usually denoted by verbal (e.g. \textit{to ride}) or nominal expressions (e.g. \textit{a discussion}). The temporal structure of an event is partly inherent to such expressions, i.e.\ lexical aspect \citep[or \textit{Aktionsart};][]{vendler1967linguistics}, and partly determined by their textual context. The events mentioned in a text are connected by (asymmetrical) temporal relations \cite{pustejovsky2003timeml}, which can be expressed explicitly (e.g., through adverbial markers such as \textit{after} or \textit{before}) or implicitly.

In particular, we focus on 3 aspects of temporal reasoning illustrated with examples from \cref{fig:temporal-grounding-qualitative-examples}: 

\begin{itemizesquish}{-0.3em}{0.5em}
    \item \textbf{Commonsense Knowledge about Events}: A grounded model should exhibit commonsense knowledge with respect to the typical duration, occurrence, frequency, and structure of events. For instance, when presented with the question ``\textit{How long has Safti been in love with Edwina?}'', the model should recognise that ``\textit{for years}'' is a more probable answer than ``\textit{10 seconds}'', due to the typical duration of this emotional state.
    \item  \textbf{Ordering Events along a Timeline}: Such a model should also be able to infer the chronology of events from narratives (e.g., news) where they are not presented in temporal order. For example, the model should deduce that ``\textit{Tim drank a little too much.}'' occurs before that ``\textit{His golf game was awful.}'', due to the causal relation between these two events.
    \item \textbf{Satisfying Temporal Constraints}: A grounded model should maintain \textit{self-consistency} in its predictions. One important constraint is that contradicting timelines cannot co-exist. If, with respect to a certain context, the model classifies ``\textit{The Obama administration joined the West before June 12}'' as true, then it must classify the opposite statement ``\textit{The Obama administration joined the West after June 12}'' as false.
\end{itemizesquish}

\section{Temporal Grounding Evaluation}

We probe the temporal grounding of LLMs from the GPT and LLaMA families, which collectively represent a wide range of LLM types. 
The models we tested can be grouped into two categories: 1) foundation models (e.g., LLaMA 1 and 2; \citealt{touvron2023llama, touvron2023llama2}) pre-trained on unlabelled texts, and 2) LLMs fine-tuned on instruction-following (e.g., Alpaca and \texttt{text-davinci-002}) or conversational objectives (e.g., GPT-4; \citealt{openai2023gpt4} and LLaMA-2-chat; \citealt{touvron2023llama2}). We use the OpenAI official API\footnote{\url{https://platform.openai.com/}} and the \texttt{transformers} library \cite{wolf-etal-2020-transformers} to conduct all experiments. 

\subsection{Benchmarks for Temporal Grounding}
\label{sec:temporal-grounding-benchmarks}
For the evaluation, we choose two off-the-shelf benchmarks, McTACO \cite{zhou-etal-2019-mctaco-dataset} and CaTeRS \cite{mostafazadeh-etal-2016-caters-dataset}, and curate a third benchmark starting from TempEvalQA \cite{llorens-etal-2015-TempEvalQA-dataset} to measure the three aspects of temporal grounding identified in \cref{sec:temporal-grounding-formalism}. Table~\ref{fig:temporal-grounding-qualitative-examples} showcases an example from each benchmark. 

\textbf{McTACO} is a benchmark for evaluating temporal commonsense knowledge through multiple choice question answering. McTACO consists of 13K tuples in the form of (context, question, candidate answers). McTACO examples fall within one of five categories: \textit{duration} (how long an event lasts), \textit{temporal ordering} (typical order of events), \textit{typical time} (when an event happens customarily), \textit{frequency} (how often an event occurs), and \textit{stationarity} (whether a state holds for a very long time or indefinitely).

\textbf{CaTeRS} is a benchmark for event ordering, consisting of 1684 instances. This task involves identifying events mentioned in a text and arranging them in chronological order. To solve this task, a model must rely on explicit clues as well as commonsense knowledge about relations between events, to reason about the underlying timeline.



\textbf{TempEvalQA-Bi}, derived from TempEvalQA \cite{llorens-etal-2015-TempEvalQA-dataset}, is our curated dataset for assessing a model's self-consistency in temporal reasoning. TempEvalQA comprises 294 human-annotated question--answer triples in the format of (context, question, yes/no answer), focusing on temporal relations between pairs of events. To construct TempEvalQA-Bi, we selectively choose question--answer pairs that involve the \textit{before}/\textit{after} temporal relation. We then swap the temporal relation to its opposite to create negative question--answer pairs. For instance, if the original pair is (``\textit{Is} E1 \textit{after} E2?'', ``\textit{yes}''), the corresponding negative pair will be (``\textit{Is} E1 \textit{before} E2?'', ``\textit{no}''), and vice versa. After this post-processing, TempEvalQA-Bi consists of 224 test examples with 448 question--answer pairs. We assess the model's performance by evaluating its predictions on both the original and negative pairs, considering the model's answer as correct only if it accurately predicts both pairs.

\begin{table*}[t]
\centering
\begin{tabular}{l|ll|ll|l|}
\toprule
& \multicolumn{4}{c|}{\textbf{McTACO}} & \textbf{CaTeRS} \\
\multicolumn{1}{c|}{\multirow{2}{*}{\textbf{Models}}} & \multicolumn{2}{c}{\textbf{Zero-shot}} & \multicolumn{2}{c|}{\textbf{Few-shot}} & \\ \cmidrule{2-6} 
\multicolumn{1}{c|}{}                                 & \textbf{Strict Acc.}  & \textbf{F1}     & \textbf{Strict Acc.} & \textbf{F1} & \textbf{Pair Acc.}   \\ \midrule
RoBERTa                                               & 43.62                 & 72.34           & \cellcolor{gray!25}                    & \cellcolor{gray!25}  & \cellcolor{gray!25}            \\
TemporalBART & \cellcolor{gray!25}                    & \cellcolor{gray!25} & \cellcolor{gray!25}                    & \cellcolor{gray!25} & \textbf{77.06} \\
Human                                                 & \textbf{75.80}         & \cellcolor{gray!25}               & \cellcolor{gray!25}                    & \cellcolor{gray!25}  & \cellcolor{gray!25}            \\ \midrule

GPT-4                                                 & \textbf{28.45}                 & 35.88           & \textbf{50.15}       & \textbf{65.27} & \textbf{60.51} \\
text-davinci-003                                      & 26.05                 & \textbf{48.30}            & 33.56                & 65.04 & 53.47          \\ \midrule

LLaMA-7B                                              & 14.39$_{2.82}$        & 35.30$_{15.18}$           & 20.17$_{2.46}$       & 22.39$_{5.07}$    & \hspace{5pt}3.76$_{4.58}$       \\
Alpaca-7B                                             & \textbf{21.75$_{5.22}$}        & \textbf{52.17$_{9.69}$}  & \textbf{30.03$_{10.11}$}      & 44.10$_{18.36}$ & 10.37$_{4.91}$           \\
LLaMA-13B                                             & 15.67$_{3.42}$        & 36.59$_{14.69}$           & 24.37$_{6.08}$       & 34.99$_{19.01}$ & \hspace{5pt}5.27$_{5.51}$          \\
LLaMA-33B                                             & 17.24$_{3.36}$        & 33.20$_{15.07}$           & 29.70$_{4.79}$       & 47.57$_{8.36}$  & 14.38$_{10.77}$         \\
LLaMA-65B                                             & 18.14$_{5.63}$        & 46.83$_{6.51}$           & 26.13$_{12.15}$      & \textbf{47.84$_{2.65}$}  & \textbf{21.02$_{10.27}$ }      \\ \midrule

LLaMA-2-7B                                            & 11.16$_{1.55}$        & \textbf{42.55$_{12.29}$}           & 21.74$_{3.83}$       & 32.94$_{17.56}$ & \hspace{5pt}5.85$_{2.06}$         \\
LLaMA-2-13B                                           & 15.69$_{3.49}$        & 39.35$_{15.55}$           & \textbf{29.75$_{0.69}$}       & \textbf{43.21$_{2.51}$} & 16.26$_{5.75}$        \\
LLaMA-2-70B                                           & 19.12$_{3.58}$        & 33.51$_{9.75}$           & 27.77$_{2.35}$       & 37.20$_{3.71}$   & 21.61$_{8.39}$      \\
LLaMA-2-chat-7B                                       & 20.74$_{3.45}$        & 28.73$_{4.48}$           & 23.00$_{3.56}$       & 31.50$_{10.18}$  & 26.32$_{2.09}$        \\
LLaMA-2-chat-13B                                      & \textbf{22.22$_{0.13}$}        & 31.67$_{9.38}$           & 28.90$_{1.04}$       & 41.63$_{5.97}$ &   30.27$_{3.02}$        \\
LLaMA-2-chat-70B                                      & 20.84$_{2.08}$        & 26.42$_{5.98}$           & 27.18$_{4.9}$        & 34.37$_{7.75}$   & \textbf{30.55$_{21.87}$}       \\ \bottomrule
\end{tabular}
\caption{Average model performance (standard deviations as subscripts). Left: McTACO for evaluating temporal commonsense reasoning in LLMs. \textbf{Strict Acc.} and \textbf{F1} follow the definitions in \citet{zhou-etal-2019-mctaco-dataset}. Right: CaTeRS results for few-shot prompting. \textbf{Pair Acc.} stands for pairwise accuracy.
In general, $(\uparrow)$ indicates that a higher value is better. The results for three prompt templates are shown in \cref{tab:mctaco-results-all-templates} for McTACO and in \cref{tab:caters-all-prompts} for CaTeRS.}
\label{tab:mctaco-results}
\end{table*}


\subsection{Inference of LLMs}
We now provide details of performing inference with LLMs on these three benchmarks.

\noindent \textbf{Multiple-choice Question Answering} is used for benchmarking LLMs' performance on McTACO. For GPT models, which do not always provide access to output probabilities, we simply provide the context, question, and \textit{all} the candidate answers in a single prompt and generate one or more answers. 
For LLaMA models, a more robust setup consists in 
presenting the context, question, and \textit{each} of candidate answers separately in multiple prompts. We calculate the likelihood of decoding ``\textit{True}'' and ``\textit{False}'' and assign the label with the higher value to each candidate answer.

\noindent \textbf{Sequence-to-Sequence Generation.} We follow \citet{lin-etal-2021-TemporalBART} to formulate the evaluation for CaTeRS as a sequence-to-sequence problem. The model takes the event series as the input and temporally sorts it as the output. We only test prompting with in-context learning for CaTeRS and not zero-shot inference because sorting the unordered events requires the model to also generate and sort special event tokens, which are unseen during pretraining. 

\noindent \textbf{Yes/No Question Answering}, in the TempEvalQA-Bi benchmark, requires the model to predict either Yes or No as the answer for each question.
To probe the models' predictions, we use 1) greedy decoding for GPT models, where we generate the token with the maximum likelihood to complete a given prompt. We then map all possible verbal tokens (e.g., yes/no, true/false and 1/0) into binary form. 2) Likelihood-based evaluation for LLaMA models: similar to \citet{ruis2022largelangaugemodelNotZeroshotCommunicator} an \citet{liu2023LanguageModeldosenotmodelAmbiguity}, we append ``\textit{yes}'' and ``\textit{no}'' as the answer in our prompt template. We then estimate the likelihood of both with LLMs. The label with the higher likelihood value is taken as the model's prediction.

\section{Results and Discussion}
\label{sec:results}

In this section, we describe our main results on three datasets: McTACO, CaTeRS,
TempEvalQA-Bi. We also analyse LLMs' sensitivity to prompting in temporal reasoning, scaling behaviour, and the impact of chain-of-thought prompting.

\subsection{Temporal Commonsense: McTACO}
\begin{figure*}
    \centering
    \includegraphics[width=\textwidth]{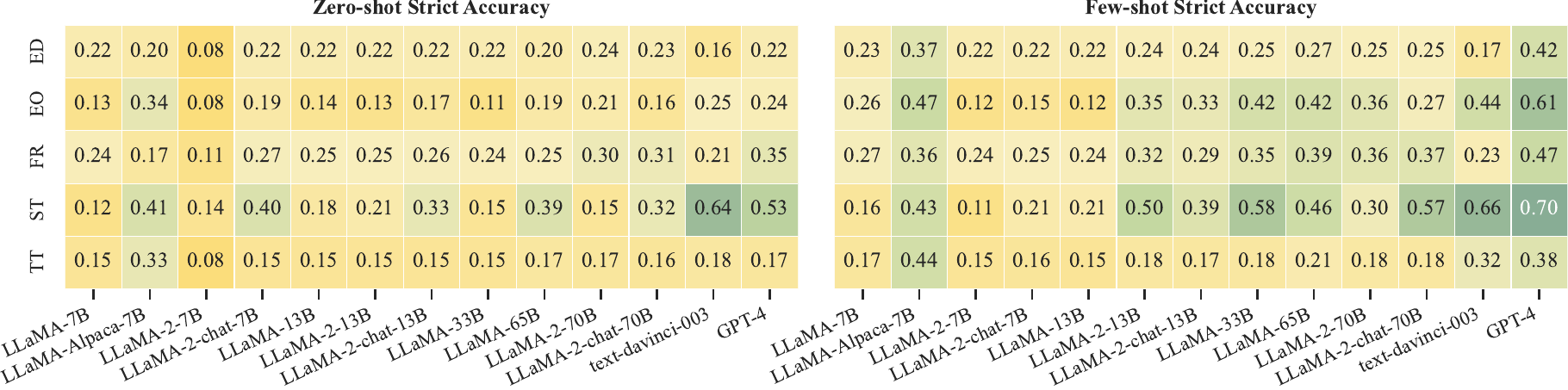}
    \caption{Strict accuracy of all tested LLMs for different reasoning categories in McTACO. The rows represent all reasoning categories in McTACO, which are Event Duration (ED), Event Ordering (EO), Frequency (FR), Stationary (ST) and Typical Time (TT). The columns represent several LLMs. }
    \label{fig:mctaco-categorical-accuracy}
\end{figure*}

\noindent \textbf{Evaluation Settings.} 
First, we probe temporal commonsense knowledge.
Table~\ref{tab:mctaco-results} shows each model's strict accuracy and F1 score on McTACO. Following \citet{zhou-etal-2019-mctaco-dataset}, strict accuracy is the percentage of questions where the model predicts \textit{all} the correct candidate answers, whereas the F1 score is calculated from all question--answer pairs. Figure~\ref{fig:mctaco-categorical-accuracy} shows the strict accuracy split by question category.

\noindent \textbf{Discussion.} In Table~\ref{tab:mctaco-results}, GPT-4 outperforms all tested LLMs, confirming its state-of-the-art ability in temporal commonsense knowledge among LLMs. 
Within the LLaMA family, LLaMA-2-chat-13B and Alpaca-7B achieve the best performance in zero-shot and few-shot experiments, respectively. 
Alpaca-7B is comparable to LLaMA-2-chat-13B also in the zero-shot setting. We attribute its remarkable performance to its in-domain temporal tasks during instruction tuning (for more discussion, see \cref{sec:temporal-task-in-instruction-tuning}).
We also notice that all LLaMA-2-chat models considerably improve their pre-trained counterparts, indicating the importance of advanced tuning (e.g., conversational-style tuning and reinforcement learning with human feedback, RLHF; \citealt{touvron2023llama2}) in addition to pre-training. 
However, even when prompting with in-context learning, the top-performing GPT-4 lags behind the fine-tuned RoBERTa baseline in terms of F1 score and human performance in terms of strict accuracy. This finding highlights the shortcomings of state-of-the-art LLMs in learning temporal commonsense knowledge. Inspecting the performance by question category, plotted as a heatmap in Figure~\ref{fig:mctaco-categorical-accuracy}, reveals that questions requiring commonsense knowledge of Typical Time and Event Duration are the most challenging. Here, GPT-4 achieves a success rate of only around 40\%. On the other hand, questions related to Stationarity are relatively easier, with most LLMs showing adequate performance using in-context learning.

\subsection{Event Ordering: CaTeRS}
\label{sec:CaTeRS-timeline-reasoning}

We evaluate the ability of LLMs to order events on CaTeRS: results are displayed in \cref{tab:mctaco-results}. 

\noindent \textbf{Evaluation Settings.} We follow \citet{lin-etal-2021-TemporalBART} in using pairwise accuracy as a metric. This measures how many pairs of events in the output are ordered correctly by a model. This metric validates each model in two respects: 1) the model is asked to extract the events from the given text correctly. And 2) the model must order the given set of events according to their chronology, producing a semi-structured output. We also include TemporalBART from \citet{lin-etal-2021-TemporalBART} as a baseline. TemporalBART is a BART-Large model fine-tuned on re-constructing manually corrupted event series.  

\begin{figure*}[t]
     \centering
     \begin{subfigure}[b]{.7\linewidth}
         \centering
         \includegraphics[width=\linewidth]{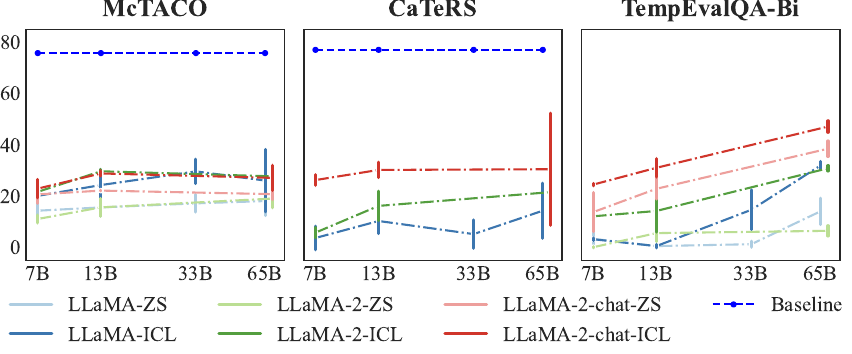}
         \caption{Scaling the model parameters.}
         \label{fig:params-scaling-all}
         \vspace{1mm}
     \end{subfigure}
     
     \begin{subfigure}[b]{.7\linewidth}
         \centering
         \includegraphics[width=\linewidth]{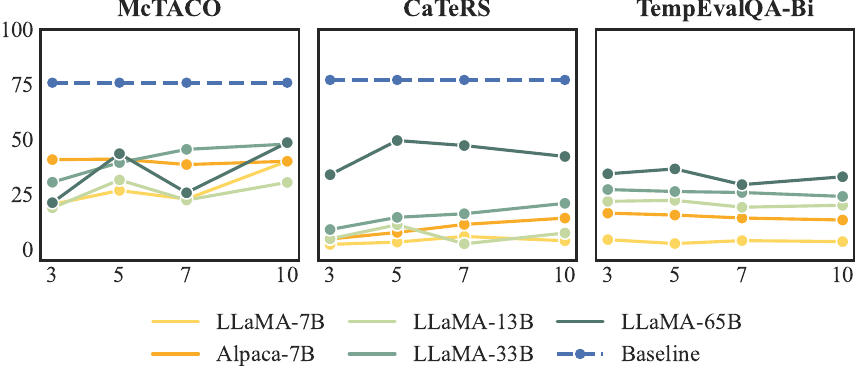}
         \caption{Scaling the examples of in-context learning.}
         \label{fig:k-scaling-all}
     \end{subfigure}
     \caption{The performance curve for scaling experiments. We report the strict accuracy for McTACO, pairwise accuracy for CaTeRS and accuracy for TempEvalQA-Bi. (a): The error bars show the standard deviation over three prompt templates. (b): The baseline for McTACO is Human, and for CaTeRS is TemporalBART.}
     \label{fig:scaling-effect}
\end{figure*}

\noindent \textbf{Discussion.} First, we confirm again the positive gains from conversational tuning and RLHF by observing the large advantage of LLaMA-2-chat models over LLaMA-2 models. 
However, we do not observe improvements by increasing the number of in-context learning examples (see Figure~\ref{fig:k-scaling-all}) or model parameters (see Figure~\ref{fig:params-scaling-all}). 
The proprietary models, \texttt{text-davinci-003} and GPT-4,  achieve higher pairwise accuracy than all LLaMA models; however, they lag behind a small fine-tuned model, TemporalBART \cite{lin-etal-2021-TemporalBART}, by a large margin. 

\subsection{Satisfaction of Temporal Constraints: TempEvalQA-Bi}
\label{sec:tempeval-bi-results}

In Table~\ref{tab:tempevalbi-result}, we report the performance of LLMs on our curated bi-directional TempEvalQA-Bi dataset. 

\noindent \textbf{Evaluation Metrics.} We evaluate the models in terms of their accuracy (\textbf{Acc.}), where the model is required to correctly predict a yes/no answer for a question with a \textit{before}/\textit{after} relation (see Table~\ref{fig:temporal-grounding-qualitative-examples} for an example). We also include a metric, namely the percentage of inconsistent predictions (\textbf{Inc.}), to shed further light on the model behaviour. This measures the number of times a model predicts the \textit{same} yes/no label for both the original question and a version where the temporal relation is flipped (everything else remaining the same). 
 For example, if models predict \textit{yes} for a question with a \textit{before} relation, ideally they should predict \textit{no} for the corresponding question with \textit{after}.
We can compare the two metrics, \textbf{Acc.}\ and \textbf{Inc.}, to probe the model's ability to answer correctly and consistently, respectively.

\noindent \textbf{Discussion} We first observe that most models perform poorly. The best model in the LLaMA family (LLaMA-2-chat-70B with in-context learning) can only solve 46\% of questions correctly. Moreover, if we flip the temporal relation of the question, most models fail to convert their output predictions accordingly, resulting in significant percentages of inconsistent predictions. This observation suggests that inconsistent predictions are the main cause of low performance on this task.

On the other hand, solid improvements result from advanced tuning, including instruction/conversation tuning and RLHF, enabling the model's predictions to be more consistent. The Alpaca-7B model dramatically reduces the rate of inconsistencies and improves accuracy over LLaMA-7B. The same holds true for LLaMA-2 and LLaMA-2-chat, \texttt{text-davinci} and \texttt{text-davinci-002}/\texttt{002}. In-context learning can also be beneficial, but it yields mixed results for some models, such as LLaMA-7B and 13B. 

Finally, GPT-4 again achieves state-of-the-art accuracy by solving around 67\% of our curated questions. Although it significantly reduces the percentage of inconsistent predictions compared to other models, it still predicts the same answer for around 27\% of the questions and their flipped versions.

\begin{table*}[t]
\centering
\begin{tabular}{l|ll|ll}
\toprule
\multicolumn{1}{c|}{\multirow{2}{*}{\textbf{Models}}} & \multicolumn{2}{c|}{\textbf{Zero-shot}} & \multicolumn{2}{c}{\textbf{Few-shot}} \\ \cmidrule{2-5} 
\multicolumn{1}{c|}{}                                 & \textbf{Acc.} ($\uparrow$)    & \textbf{Inc.} ($\downarrow$)    & \textbf{Acc.} ($\uparrow$)   & \textbf{Inc.} ($\downarrow$)  \\ \midrule
GPT-4                                                 & \textbf{64.29}            & \textbf{31.25}                & \textbf{67.41}           & \textbf{27.23}               \\
text-davinci-003                                      & 27.68            & 69.64                & 33.93           & 62.05               \\
text-davinci-002                                      & 16.52            & 77.83                & 36.16           & 60.71               \\
davinci                                               & 14.73            & 79.02                & 13.39           & 79.91               \\ \midrule
LLaMA-7B                                              & \hspace{5pt}3.42$_{2.46}$             & 94.79$_{3.64}$                & \hspace{5pt}3.27$_{0.51}$            & 94.94$_{1.57}$               \\
LLaMA-Alpaca-7B                                       & 10.12$_{2.29}$            & 83.63$_{5.08}$                & 13.10$_{6.00}$           & 77.23$_{7.37}$               \\
LLaMA-13B                                             & \hspace{5pt}0.60$_{0.68}$             & 97.77$_{3.49}$                & \hspace{5pt}0.60$_{0.68}$            & 99.25$_{0.51}$               \\
LLaMA-33B                                             & \hspace{5pt}1.34$_{1.34}$             & 98.22$_{1.18}$                & 14.73$_{7.74}$           & 83.33$_{9.43}$               \\
LLaMA-65B                                             & \textbf{14.14$_{5.17}$}            & \textbf{83.48$_{6.14}$}                & \textbf{31.99$_{1.57}$}           & \textbf{60.42$_{4.47}$}               \\ \midrule
LLaMA-2-7B                                            & \hspace{5pt}0.15$_{0.26}$             & 99.85$_{0.26}$                & 11.90$_{0.52}$           & 85.12$_{2.62}$               \\
LLaMA-2-13B                                           & \hspace{5pt}5.65$_{3.3}$              & 92.86$_{3.81}$                & 13.69$_{7.63}$           & 83.63$_{8.00}$               \\
LLaMA-2-70B                                           & \hspace{5pt}6.55$_{2.01}$             & 92.41$_{3.13}$                & 29.76$_{2.73}$           & 65.77$_{2.02}$               \\
LLaMA-2-chat-7B                                       & 13.84$_{7.63}$            & 83.33$_{7.82}$                & 23.51$_{2.20}$           & 70.09$_{0.77}$               \\
LLaMA-2-chat-13B                                      & 22.92$_{4.03}$            & 72.91$_{5.58}$                & 31.69$_{3.22}$           & 62.95$_{3.57}$               \\
LLaMA-2-chat-70B                                      & \textbf{38.54$_{3.04}$}            & \textbf{58.03$_{2.36}$}                & \textbf{46.42$_{1.18}$}           & \textbf{48.96$_{2.01}$}               \\ \bottomrule
\end{tabular}
\caption{Average model performance (standard deviations as subscripts) evaluated on our curated bi-directional TempEvalQA benchmark. \textbf{Acc.} and \textbf{Inc.} stand for accuracy and the percentage of inconsistent predictions. $(\uparrow) / (\downarrow)$ indicate that higher / lower values are better, respectively. 
The detailed results for three prompt templates are presented in \cref{tab:tempevalbi-result-all-prompts}. }
\label{tab:tempevalbi-result}
\end{table*}

\subsection{Sensitivity to Different Prompts}

Large language models have been shown to be sensitive to the wording of their prompts \cite{webson-pavlick-2022-prompt, reynolds2021prompt-programming, tan-etal-2022-msp, ruis2022largelangaugemodelNotZeroshotCommunicator,liu2023LanguageModeldosenotmodelAmbiguity}. Similar to \citet{ruis2022largelangaugemodelNotZeroshotCommunicator} and \citet{liu2023LanguageModeldosenotmodelAmbiguity}, to test the randomness introduced by this sensitivity in our experiments, we manually curate three different prompt templates and measure the variation of performance across these different wordings for the LLaMA family. 

We experiment with three different kinds of prompt templates: 1) \textit{semi-structured prompt}: we indicate the instruction, context, and pairs of input and output (for in-context learning) in a sequence, separated by special symbols (e.g., new-line). 2) \textit{Natural prompt}: we interleave the context and input--output pairs in the instruction, to make the template closer to natural language. 3) \textit{Text continuation prompt}: given that LLaMA is not trained with instruction-following tasks, we introduce a text-continuation style prompting where we only ask the model to complete the input. We provide all our prompt templates in Table~\ref{tab:prompt-templates}.

We report the standard deviation of model performance in the result tables for each of the three datasets. Standard deviations range from 0.13 for the LLaMA-2-chat-13B model to 5.63 for LLaMA-65B when considering all templates in McTACO in zero-shot prompting. The variation caused by prompt templates is slightly higher for zero-shot results in TempEvalQA-Bi, where the standard deviation ranges from 0.26 for LLaMA-2-7B to 7.63 for LLaMA-2-chat-7B. 
The few-shot experiments tend to be more varied, because of the increased length of each prompt due to adding additional exemplars. 
Overall, these findings show that the prompt wording does not affect the main conclusions of our experiments. 
However, they also confirm that LLMs---especially large-scale ones---are brittle with respect to prompt wording.

\subsection{Scaling Behaviour}
\label{sec:scaling-effects}

\noindent \textbf{Scaling the Model Parameters.} We perform scaling experiments (refer to Figure~\ref{fig:params-scaling-all}) to explore the influence of parameter increments on the performance of LLMs. 
We exclusively execute this ablation within the LLaMA family due to the unavailability of information on model sizes for the proprietary GPT models.
The largest 70B LLaMA-2-chat model can generally achieve the best performance on our three datasets. 
However, for many models, performance is only weakly correlated with their size. For example, the LLaMA-13B model, rather than the smaller LLaMA-7B one, performs the worst on both McTACO and TempEvalQA-Bi. 
We also observe a general trend such that LLM performance tends to saturate and cannot further increase after the parameter size exceeds 13B on McTACO and CaTeRS.
Finally, comparing the LLMs with the much smaller baselines of TemporalBART and RoBERTa shows that model specialisation is more impactful than an increase in scale for temporal grounding.  

\noindent \textbf{Scaling the Few-shot Examples.} We begin by conducting scaling experiments (please refer to Figure~\ref{fig:k-scaling-all}) to investigate the impact of the number of few-shot examples ($k$) on model performance. We focus our experiments on the LLaMA family. The results on McTACO and CaTeRS indicate that increasing $k$ generally leads to a slight performance improvement, albeit minimal. Conversely, we observe a slight decrease in performance on TempEvalQA-Bi when increasing $k$. This decrease can be attributed to our choice of increasing only the number of question--answer pairs for the same context, as using multiple contexts would surpass the model's maximum input limit. Notably, McTACO and CaTeRS exhibit true scaling with few-shot examples, supporting the idea that examples must be diverse to contribute incrementally to performance
\cite{min-etal-2022-rethinking}. 



\subsection{Chain-of-thought Prompting}
\label{sec:chain-of-thought}

Finally, we investigate the impact of chain-of-thought prompting \cite{wei2022chain} on model performance in TempEvalQA-Bi, a challenging reasoning task. We apply chain-of-thought prompting to LLaMA\footnote{Since LLaMA-2 models support twice the input length of LLaMA when prompting all LLaMA-2 models, we regard the maximum input length as a hyperparameter. 
} and GPT-4 models in combination with in-context learning. The prompts we constructed are listed in Appendix~\ref{appendix:cot_examples}. \Cref{fig:chain-of-thought-results} shows that chain-of-thought prompting reduces inconsistent predictions for all models; however, the improvement of reasoning consistency does not always translate into increased accuracy. For example, we observe a drop in accuracy for LLaMA-7B, LLaMA-2-70B and LLaMA-2-70B-chat models. 
We also notice that the improvement in the state-of-the-art GPT-4 is marginal. These results indicate that current training and inference paradigms may struggle with complex temporal reasoning tasks requiring self-consistency.

\begin{figure*}[t]
    \centering
    \includegraphics[width=.8\linewidth]{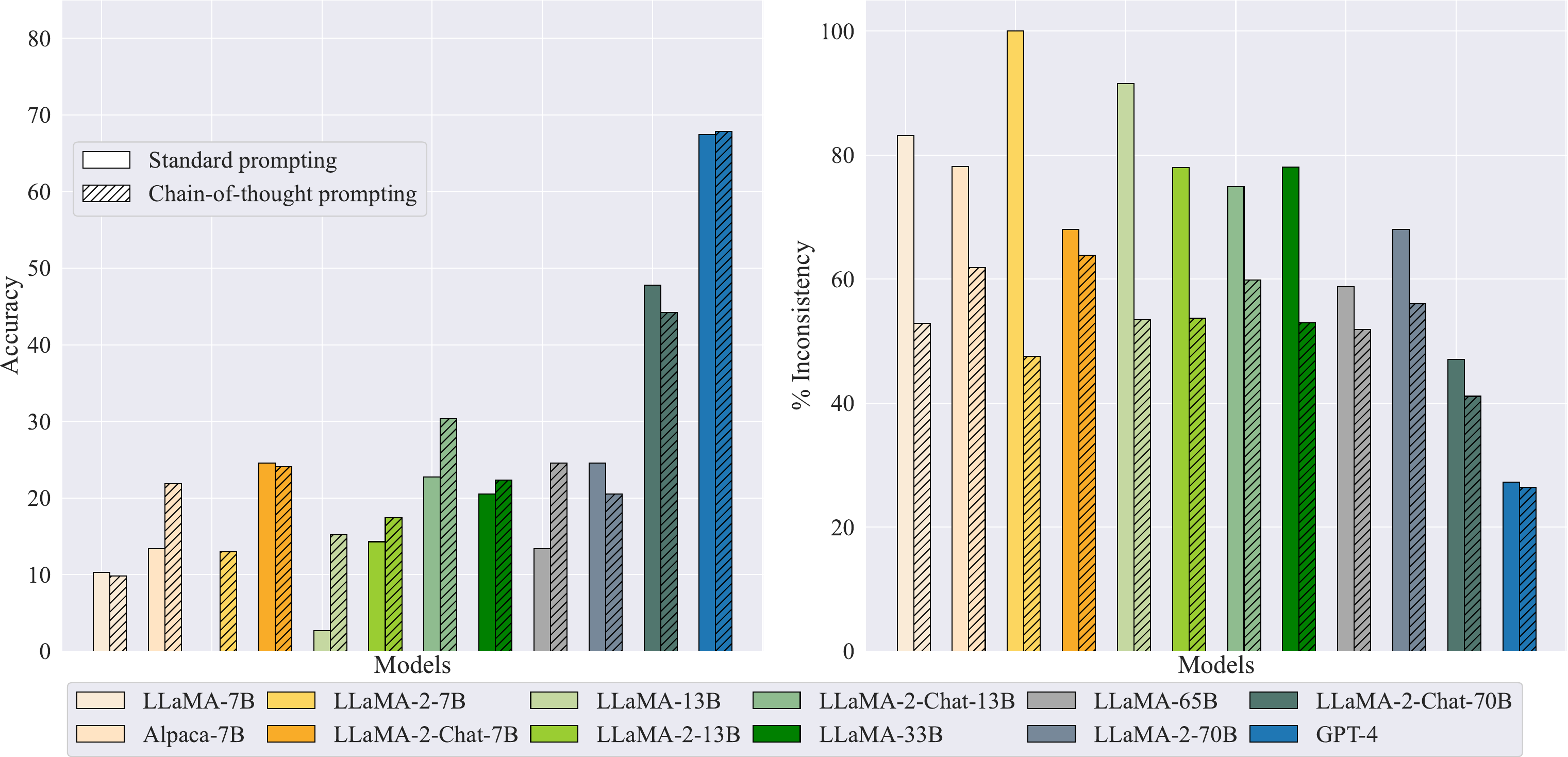}
    \caption{Accuracy (higher is better) and percentage of in-consistent reasoning (lower is better) on TempEvalQA-Bi for LLMs prompted with a chain of thoughts or not.}
    \label{fig:chain-of-thought-results}
\end{figure*}


\section{Sources of Temporal Information}
\label{sec:source-of-temporal-info}

In this section, we examine two key aspects: firstly, the extent of temporal information provided by pre-training, and secondly, the significance of supervised examples in addressing information gaps within the pre-trained model.

\subsection{How Much Temporal Information Does Pre-training Provide?}



\begin{figure}[t]
    \centering
    \includegraphics[width=\linewidth]{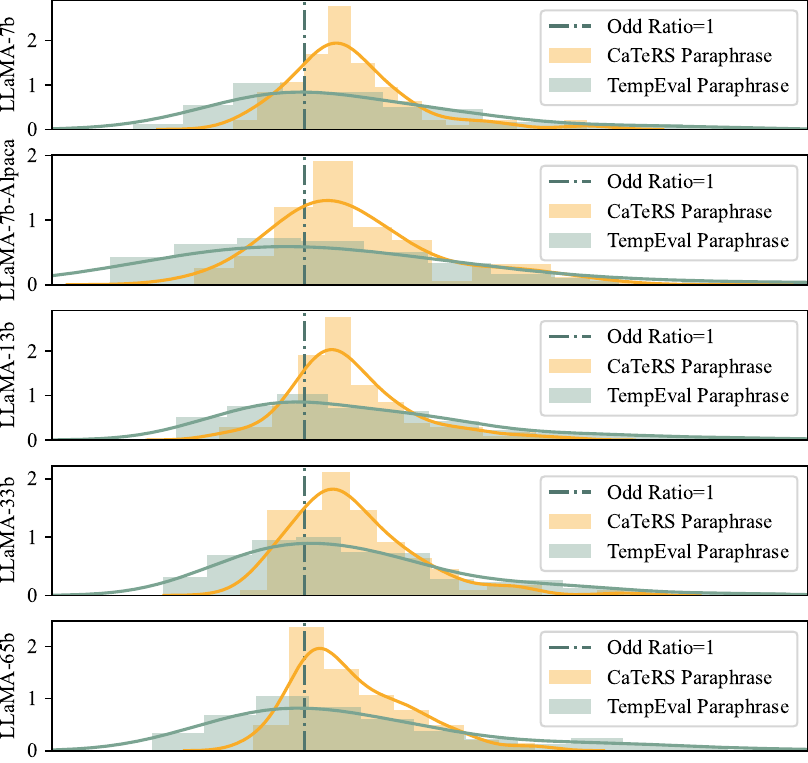}
    \caption{Density plot of the odds ratio under several LLMs (rows) for differently ordered paraphrases in CaTeRS (\hlred{orange}) and TempEvalQA-Bi (\hlgreen{green}).}
    \label{fig:odd-ratio-figure}
\end{figure}
Overall, the results from \cref{sec:results} indicate that LLMs are not sufficiently temporally grounded. We speculate that this is due to not being exposed to temporal information during their training pipeline. To verify this, we first focus on the pre-training stage.
We investigate whether the order in which events are presented in human-written texts provides a clue about their actual chronology. We measure the correlation between the temporal relations (i.e., \textit{before}/\textit{after}) annotated in TempEvalQA's training articles and the textual order of the events they refer to. This reveals that only 55.98\% of event pairs (1866 out of 3333) occur in the text in accordance with their temporal order. The Matthews correlation coefficient of 0.09 indicates a weak correlation \cite{matthews1975comparison, davenport1991phi}. 

We then study whether, as a result of this property of the data, LLMs prefer temporally ordered or unordered descriptions of the same event series, by comparing their log-likelihoods. We randomly sample 100 instances from the CaTeRS testing set and manually curate two paraphrases for each ground-truth sequence of events. For the first paraphrase, we keep the same sentence order as the event temporal order. For the second paraphrase, we permute the order of the sentences. All the resulting paraphrases are manually adjusted to be coherent and grammatically correct to exclude confounding factors. We present one example:

\noindent\fbox{%
    \parbox{\linewidth}{%
\textbf{Ground-truth Order}: [E1] \textit{Mary was very stressed about the job.} [E2] \textit{Mary got the job.}
    
\textbf{Ordered Paraphrase}: \textit{Mary was very stressed about the job, but she got it.}

\noindent
\textbf{Inverted Paraphrase}: \textit{Mary got the job. She was very stressed about it.}%
}
}

We then create another subset of paraphrases from TempEvalQA-Bi, retaining explicit temporal relation markers such as \textit{before} and \textit{after}.

We plot the distribution of the length-normalised odds ratios between corresponding paraphrases in Figure~\ref{fig:odd-ratio-figure}. Firstly, we find that the odds ratio is significantly higher than 1 on average for the CaTeRS paraphrases. This means that the probability of temporally ordered sequences is only slightly higher than their permuted counterparts. Thus, foundation models have a mild bias in preferring sequences of sentences that follow the temporal order of events. For the examples with explicit temporal relation markers (i.e., \textit{before} and \textit{after}) from TempEvalQA-Bi, the odds ratio falls around 1. Thus, the model's preference for temporally ordered and unordered sequences tends to be equalised. These results demonstrate that it may be difficult for foundation models to identify the ground-truth event ordering based on their pre-training information, as this provides a weak signal about the temporal dynamics of real-world events.
 
\subsection{Importance of Supervised Examples}
\label{sec:temporal-task-in-instruction-tuning}

Next, we investigate whether subsequent fine-tuning might compensate for the lack of temporal information during pre-training.
The current open-sourced instruction-tuning datasets commonly used to fine-tune LLMs indeed tend to include some temporal tasks. For example, Super-Natural Instruction \cite{naturalinstructions,supernaturalinstructions}, one of the most comprehensive massively multi-task benchmarks, has only 2 temporal reasoning tasks in its training set out of 756 in its \texttt{default} split.\footnote{\texttt{task1507\_boolean\_temporal\_reasoning}, \texttt{task389\_torque\_generate\_temporal\_question}.} In Table~\ref{tab:mctaco-results} and Table~\ref{tab:tempevalbi-result}, we observe Alpaca-7B consistently outperforming LLaMA-7B and being almost comparable with LLaMA-33B. In addition, we found that small-scale models fine-tuned on many in-task examples, such as RoBERTa and TemporalBART in \cref{tab:mctaco-results}, often surpass or are competitive with large-scale, state-of-the-art LLMs such as GPT-4. This lends hope that the gap with human performance can be partly filled with sufficient supervision; however, there remains an insurmountable barrier that even providing more examples cannot overcome: GPT-4 performance plateaus even increasing in-context examples (see \cref{fig:scaling-effect}). Even chain-of-thought prompting does not increase its self-consistency (see \cref{fig:chain-of-thought-results}). We hypothesise that only equipping LLMs with perception and action in a simulated or physical environment might further ground them sufficiently to enhance their temporal reasoning.

\section{Related Work}


Recently, a budding interest has emerged in probing whether large language models are grounded \citep{yun-etal-2021-vision-language-model-grounding,abdou-etal-2021-language-model-color-grounding,carta2023grounding,mahowald2023dissociating, chandu-etal-2021-grounding} with respect various properties. \citet{abdou-etal-2021-language-model-color-grounding} found a correspondence between physical colour space and the embeddings of colour terms. \citeauthor{ebert-etal-2022-trajectories} discovered that object trajectories in an environment correlate with the embeddings of motion verbs. In contrast, \citet{yun-etal-2021-vision-language-model-grounding} showed that vision-and-language pre-training fails to further ground lexical representations. Improving the grounding of LLMs is also an active area of research. 
Integrating the modelling of subject--object interactions enables better grounding during training \citep{merullo-etal-2022-pretraining, carta2023grounding}. \citet{liu2022mindEye} uses a computational physics engine as a simulator for improving how language models reason about the physical world.

\section{Conclusions}
In this study, we examined to what extent large language models (LLMs) are temporally grounded: given a textual narrative describing a series of events, we propose a framework to probe their temporal reasoning capabilities. We consider a series of tasks including commonsense knowledge about the events, ordering events along a timeline, and satisfaction of temporal constraints (which ensures the self-consistency of the LLM's internal world model).
Through a systematic evaluation, we find that state-of-the-art models such as GPT-4 still fall short of human performance and even small-scale, specialised models.
Additionally, we observed that increasing the model scale or the number of in-context examples does not necessarily result in meaningful gains. Instruction-tuning and chain-of-thought prompting are more beneficial for accuracy and self-consistency but yield diminishing returns with scale.
To explain these patterns, 
we highlight the limitations of pre-training and instruction tuning in achieving satisfactory temporal grounding and suggest potential future directions. These include exploring new training paradigms, such as leveraging action and perception in physical or simulated environments. 

\section*{Limitations}

Although we carefully select human-annotated benchmarks for evaluation, there are still some known issues. Questions in McTACO \cite{zhou-etal-2019-mctaco-dataset} may contain duplicate candidates, and some of them have all true or false candidates. Models that have not been trained on in-domain data may not be robust to this noise in evaluation. CaTeRS \cite{mostafazadeh-etal-2016-caters-dataset} contains a small proportion of highly similar or duplicated instances. A high-quality benchmark for systematically evaluating the recent model's temporal grounding ability is still essential but missing.

It is not clear whether the training of GPT models is contaminated with data leakage from these datasets. This fact may pose a potential bias in overestimating the advantage of GPT models.
Additionally, we confirm that McTACO annotation is not included in the instruction tuning of LLaMA-Alpaca; however, other tasks have texts sourced from the same corpus as McTACO. Hence, the McTACO's evaluation could also be biased to favour the Alpaca model. 



\bibliography{custom}
\bibliographystyle{acl_natbib}

\newpage
\appendix

\section{All Prompt Templates}
\label{appendix:all_prompt_templates}
We provide the three prompt templates we design for our experiments in Table~\ref{tab:prompt-templates}.

\begin{table*}[t!]
\begin{tabular}{llp{13cm}}
\toprule 
\multirow{13}{*}{\rotatebox{90}{McTACO}}  & Structured & Answer the following multiple-choice question with candidate answers according to the given passage. There can be multiple correct answers.

Passage: \{passage\}

Question:   \{question\}

Candidate answers: \{candidates\} 

The answer is \\ \cmidrule{2-3}
                         & Natural & Based on the information presented in the passage ``\{passage\}'', answer the multiple-choice question ``\{question\}'' with following candidate answers ``\{candidate\}''. There can be multiple correct answers. The correct answer(s) is/are:                                      \\ \cmidrule{2-3}
                         & Text continuation  & Finish the following text: 
                         
                         According to the passage   ``\{passage\}'', the correct answer(s) to the multiple-choice question   ``\{question\}'' with following candidate answers   ``\{candidates\}'' is/are:                                                        \\ \midrule
\multirow{10}{*}{\rotatebox{90}{CaTeRS}}  & Structured & Following the given template to order the events according to   temporality: 

\textit{[Input--Output Templates]}

Input: \{source\}

Output:                                                                                                                                         \\ \cmidrule{2-3}
                         & Natural  &  Based on temporality in the given events ``\{source\}'', arrange the events in temporal order. The order is:                                                                                                                                                                                                                                                                                \\\cmidrule{2-3}
                         & Text continuation  &   Finish the following text: 
                         
                         According to the temporality in the given events ``\{source\}'', the temporal order of the events is:                                                                                                                                                                                                                                                                          \\ \midrule
\multirow{10}{*}{\rotatebox{90}{TempEvalQA-Bi}} & Structured & Answer the question according to the article. Only answer yes or no.   

Article: \{events\}. 

Question: \{question\}. 

The answer is:                                                                                           \\ \cmidrule{2-3}
                         & Natural & Based on the information presented in the article ``\{events\}'',   answer the question ``\{question\}'' with yes or no. The answer is:                                                                                           \\ \cmidrule{2-3}
                         & Text continuation  & Finish the following text: 
                         
                         Article: \{events\}"
                         
                         The answer to the   yes or no question ``\{question\}'' according to the the article is:                                                        \\ \bottomrule
\end{tabular}
\caption{The three templates, i.e., Structured prompt, Natural prompt, Text continuation prompt, used for LLMs' inference in each dataset.}
\label{tab:prompt-templates}
\end{table*}

\section{Detailed Results for All Prompt Templates}
\label{appendix:detailed_results_for_all_prompts}

In Table~\ref{tab:tempevalbi-result-all-prompts}, we present the complete results based on all three prompts on the TempEvalQA-bi dataset. Additionally, comprehensive results utilising all three distinct prompts for McTACO are provided in Table~\ref{tab:mctaco-results-all-templates}. We only list the likelihood-based evaluation results for LLaMA models in Table~\ref{tab:mctaco-results} for simplicity, but we observe the same trend for both decoding-based and likelihood-based evaluation.
Fially, detailed results using all three different prompts for CaTeRS can be found in Table~\ref{tab:caters-all-prompts}.

\begin{table*}[t]
\centering
\scalebox{0.6}{
\begin{tabular}{ccc|cccc|cccc|cccc}
\toprule
\multirow{3}{*}{} & \multicolumn{2}{c|}{\multirow{3}{*}{\textbf{Models}}} & \multicolumn{4}{c}{\textbf{Prompt 1}} & \multicolumn{4}{c}{\textbf{Prompt 2}} & \multicolumn{4}{c}{\textbf{Prompt 3}} \\ \cmidrule{4-15}
& \multicolumn{2}{c|}{} & \multicolumn{2}{c}{\textbf{Likelihood}}     & \multicolumn{2}{c}{\textbf{Decoding}} & \multicolumn{2}{c}{\textbf{Likelihood}}  & \multicolumn{2}{c}{\textbf{Decoding}} & \multicolumn{2}{c}{\textbf{Likelihood}}     & \multicolumn{2}{c}{\textbf{Decoding}}      \\ \cmidrule{4-15} 
                                   & \multicolumn{2}{c|}{}  & \textbf{Acc.  $(\uparrow)$}  & \textbf{Inc. $(\downarrow)$} & \textbf{Acc. $(\uparrow)$}  & \textbf{Inc. $(\downarrow)$}   & \textbf{Acc. $(\uparrow)$}  & \textbf{Inc. $(\downarrow)$} & \textbf{Acc. $(\uparrow)$}  & \textbf{Inc. $(\downarrow)$} &  \textbf{Acc. $(\uparrow)$}  & \textbf{Inc. $(\downarrow)$} & \textbf{Acc. $(\uparrow)$}  & \textbf{Inc. $(\downarrow)$} \\ \midrule
                                   & LLaMA-7B & ZS  & 6.25           & 90.63             & 10.71          & 82.35   
                                                        & 1.79           & 96.43             & 0.45           & 92.16         
                                                        & 2.23           & 97.32             & 0              & 60        \\
                                   &             & ICL  & 2.68           & 96.43             & 3.57           & 91.96   
                                                        & 3.57           & 93.30             & 8.93 &  82.03
                                                        & 3.57           & 95.09             & 2.23 &  86.67        \\
                                   & Alpaca-7B   & ZS   & 12.05          & 82.14             & 12.95          & 80.36   
                                                        & 7.59           & 89.29             & 1.79           & 92.04 
                                                        & 10.71          & 79.46             & 0              & 80.49         \\
                                   &             & ICL  & 17.41          & 73.66             & 11.61          & 80.80   
                                                        & 6.25           & 85.71             & 10.71 &  77.31
                                                        & 15.63          & 72.32             & 9.38 & 76.67         \\
                                   & LLaMA-13B   & ZS   & 0.45           & 99.55             & 4.02           & 90.41                       & 1.34           & 93.75             & 0 & 94.23
                                                        & 0              & 100                 & 0 &  96.15        \\
                                   &             & ICL  & 1.34           & 98.66             & 2.68           & 95.98                       & 0.45           & 99.55             & 1.34 & 85.11 
                                                        & 0              & 99.55               & 0 &  100        \\
                                   & LLaMA-33B   & ZS   & 2.68           & 96.88             & 11.16          & 86.16                       & 0.89           & 99.11             & 5.36 &  84.83
                                                        & 0.45           & 98.66             & 0 &  100        \\
                                   &             & ICL  & 19.20          & 78.57             & 17.41          & 79.46                       & 5.80           & 94.20             & 17.86 & 72.93 
                                                        & 19.20          & 77.23             & 24.55 &  69.72        \\
                                   & LLaMA-65B   & ZS   & 19.64          & 78.13             & 9.38           & 73.39                       & 13.39          & 82.14             & 4.91 &  84.34
                                                        & 9.38           & 90.18             & 0 & 100         \\
                                   &             & ICL  & 32.14 & 63.84    & 33.93 & 62.50   & 30.36 & 62.05 & 5.36 & 87.60 
                                                    & 33.48 & 55.36 & 27.68 & 58.20 \\ 
                                                    & LLaMA-2-7B       & ZS  & 0     & 100   & 0     & 100   & 0.45 & 99.55 & 0     & 1     & 0     & 100   & 0     & 100   \\
                                   &                  & ICL & 11.61 & 87.05 & 0     & 100   & 11.6  & 82.14 & 4.91  & 77.7  & 12.5  & 86.16 & 11.16 & 80.35 \\
                                   & LLaMA-2-13B      & ZS  & 9.38  & 88.84 & 10.26 & 78.05 & 3.12  & 96.43 & 0     & 97.42 & 4.46  & 93.3  & 0.45  & 76.47 \\
                                   &                  & ICL & 18.75 & 78.57 & 12.5  & 76.76 & 4.91  & 92.85 & 4.02  & 92.27 & 17.41 & 79.46 & 13.83 & 80.6  \\
                                   & LLaMA-2-70B      & ZS  & 6.7   & 92.41 & 0     & 100   & 8.48  & 89.28 & 11.16 & 78.92 & 4.46  & 95.54 & 0.9   & 44.44 \\
                                   &                  & ICL & 26.78 & 67.86 & 14.29 & 80.38 & 32.14 & 63.83 & 25    & 62.63 & 30.36 & 65.63 & 28.57 & 66.21 \\
                                   & LLaMA-2-7B-chat  & ZS  & 20.98 & 75.44 & 21.88 & 73.66 & 5.8   & 91.07 & 5.8   & 93.12 & 14.73 & 83.48 & 8.04  & 78.34 \\
                                   &                  & ICL & 20.98 & 70.98 & 19.64 & 72.94 & 24.55 & 69.64 & 20.98 & 72.65 & 25    & 69.64 & 20.98 & 68.49 \\
                                   & LLaMA-2-13B-chat & ZS  & 23.21 & 72.76 & 8.92  & 80.77 & 18.75 & 78.57 & 15.17 & 77.42 & 26.79 & 67.41 & 0.89  & 81.82 \\
                                   &                  & ICL & 30.8  & 62.94 & 21.88 & 72.64 & 29.01 & 66.52 & 23.21 & 71.95 & 35.26 & 59.38 & 30.8  & 62.5  \\
                                   & LLaMA-2-70B-chat & ZS  & 41.96 & 55.35 & 13.39 & 66.67 & 37.5  & 58.92 & 30.8  & 59.38 & 36.16 & 59.82 & 0.45  & 60    \\
                                   &                  & ICL & 47.32 & 46.88 & 37.05 & 47.64 & 46.86 & 49.1  & 47.32 & 45.7  & 45.09 & 50.89 & 47.77 & 48.66 \\ \bottomrule
\end{tabular}}
\caption{Detailed Model performance evaluated on our curated bi-directional TempEvalQA-bi benchmark with different prompt templates. \textbf{Acc.} and \textbf{Inc.} stand for accuracy and in-consistent prediction rate. $(\uparrow) / (\downarrow)$ indicate that higher / lower values are better, respectively.}
\label{tab:tempevalbi-result-all-prompts}
\end{table*}

\begin{table*}[]
\centering
\scalebox{0.9}{
\begin{tabular}{cl|cc|cc|cc}
\toprule
\multicolumn{2}{c|}{\multirow{2}{*}{\textbf{Models}}} & \multicolumn{2}{c}{\textbf{Prompt 1}} & \multicolumn{2}{c}{\textbf{Prompt 2}} & \multicolumn{2}{c}{\textbf{Prompt 3}} \\ \cmidrule{3-8} 
\multicolumn{2}{c|}{}                                 & \textbf{Strict Acc. $(\uparrow)$}   & \textbf{F1 $(\uparrow)$}  & \textbf{Strict Acc. $(\uparrow)$}   & \textbf{F1 $(\uparrow)$}  & \textbf{Strict Acc. $(\uparrow)$}   & \textbf{F1 $(\uparrow)$}  \\ \midrule
\multirow{2}{*}{LLaMA-7B}                & ZS         & 17.64                  & 19.06        & 12.69                  & 37.72        & 12.84                  & 49.12        \\
                                         & ICL        & 22.15                  & 27.56        & 20.95                  & 22.18        & 17.42                  & 17.42        \\
\multirow{2}{*}{Alpaca-7B}         & ZS         & 27.63                  & 61.19        & 17.64                  & 41.92        & 19.97                  & 53.41        \\
                                         & ICL        & 40.92                  & 62.59        & 20.95                  & 25.86        & 28.23                  & 43.84        \\
\multirow{2}{*}{LLaMA-13B}               & ZS         & 19.07                  & 20.77        & 15.69                  & 39.20        & 12.24                  & 49.80        \\
                                         & ICL        & 19.07                  & 19.40        & 23.05                  & 29.41        & 31.01                  & 56.16        \\
\multirow{2}{*}{LLaMA-33B}               & ZS         & 17.94                  & 18.24        & 20.20                  & 32.98        & 13.59                  & 48.39        \\
                                         & ICL        & 33.71                  & 45.04        & 31.01                  & 56.90        & 24.40                  & 40.78        \\
\multirow{2}{*}{LLaMA-65B}               & ZS         & 23.35                  & 51.30        & 18.92                  & 39.36        & 12.16                  & 49.84        \\
                                         & ICL        & 34.31                  & 44.83        & 31.91                  & 48.91        & 12.16                  & 49.80        \\ 
\multirow{2}{*}{LLaMA-2-7B}       & ZS  & 9.38  & 28.36 & 12.23 & 49.94 & 11.86 & 49.35 \\
                 & ICL & 17.56 & 18.22 & 25.07 & 52.38 & 22.59 & 28.23 \\
\multirow{2}{*}{LLaMA-2-13B}      & ZS  & 19.14 & 21.48 & 15.76 & 46.72 & 12.16 & 49.84 \\
                 & ICL & 30.55 & 40.32 & 29.35 & 44.66 & 29.35 & 44.66 \\
\multirow{2}{*}{LLaMA-2-70B}      & ZS  & 21.92 & 26.45 & 20.34 & 29.45 & 15.09 & 44.63 \\
                 & ICL & 28.9  & 38.38 & 29.35 & 40.18 & 25.07 & 33.05 \\
\multirow{2}{*}{LLaMA-2-7B-chat}  & ZS  & 23.72 & 29.47 & 16.96 & 32.8  & 21.54 & 23.93 \\
                 & ICL & 20.04 & 23.77 & 26.95 & 43.04 & 22    & 27.69 \\
\multirow{2}{*}{LLaMA-2-13B-chat} & ZS  & 22.07 & 25.32 & 22.29 & 27.25 & 22.29 & 42.44 \\
                 & ICL & 27.7  & 34.74 & 29.5  & 45.08 & 29.5  & 45.07 \\
\multirow{2}{*}{LLaMA-2-70B-chat} & ZS  & 23.04 & 26.98 & 18.91 & 20.18 & 20.57 & 32.11 \\
                 & ICL & 31.15 & 40.15 & 28.68 & 37.4  & 21.7  & 25.56 \\
                                         \bottomrule
\end{tabular}}
\caption{Detailed model performance on McTACO using all prompt templates. \textbf{Strict Acc.} and \textbf{F1} stand for strict accuracy and the F-1 score as in \cite{zhou-etal-2019-mctaco-dataset}.  $(\uparrow)$ indicates higher values are better.}
\label{tab:mctaco-results-all-templates}
\end{table*}

\begin{table*}[t]
\centering
\begin{tabular}{c|c|c|c|c}
\toprule
\textbf{Category}      & \textbf{Models}             & \textbf{T1 PAcc.}     & \textbf{T2 PAcc.} & \textbf{T3 PAcc.} \\ \midrule
\multirow{11}{*}{LLaMA} & LLaMA-7B                    & 8.9                  & 2.3              & 0.09             \\
                       & LLaMA-Alpaca-7B             & 13.86                & 4.75             & 12.5             \\
                       & LLaMA-13B                   & 0                    & 4.82             & 10.99            \\
                       & LLaMA-33B                   & 26.78                & 9.05             & 7.32             \\
                       & LLaMA-65B                   & 15.54                & 33.98            & 13.54            \\ 
                       & LLaMA-2-7B       & 8.15  & 4.19  & 5.21  \\
                       & LLaMA-2-13B      & 15.67 & 10.83 & 22.28 \\
                       & LLaMA-2-70B      & 20.97 & 30.3  & 13.56 \\
                       & LLaMA-2-7B-chat  & 23.99 & 28.04 & 26.93 \\
                       & LLaMA-2-13B-chat & 27.1  & 33.12 & 30.6  \\
                       & LLaMA-2-70B-chat & 17.46 & 55.79 & 18.39 \\

\bottomrule
\end{tabular}
\caption{Detailed Model performance evaluated on our curated CaTeRS benchmark with different prompt templates. \textbf{T1/2/3} and \textbf{PAcc.} stand for Template 1/2/3 and pair-wise accuracy.}
\label{tab:caters-all-prompts}
\end{table*}

\section{Chain-of-thought Reasoning Examples}
\label{appendix:cot_examples}
We provide an example of a model's input using chain-of-thought prompting on TempEvalQA-Bi in Table~\ref{tab:prompt-cot}.
Table~\ref{tab:cot-example-blog}, Table~\ref{tab:cot-example-news}, and Table~\ref{tab:cot-example-wikipedia} showcase a selection of generations obtained from LLaMA-65B and GPT-4 models using chain-of-thought prompting on the TempEvalQA-bi dataset.

\begin{table*}[t!]
\begin{tabular}{p{15.5cm}}
\toprule 
{\color{Orange} \itshape Answer the question according to the article. Give step-by-step explanations and then answer yes or no. 

Article: Tired of being sidelined, Hungarian astronaut Bertalan Farkas is leaving for the United States to start a new career, he said Saturday. \"Being 48 is too early to be retired,\" a fit-looking Farkas said on state TV's morning talk show. With American astronaut Jon McBride, Farkas set up an American-Hungarian joint venture called Orion 1980, manufacturing space-travel related technology. Farkas will move to the company's U.S. headquarters. Farkas, an air force captain, was sent into space on board the Soyuz 36 on May 26, 1980. He spent six days aboard the Salyut 6 spacecraft with three Soviet astronauts, Valery Kubasov, Leonid Popov and Valery Riumin. McBride, 54, of Lewisburg, West Virginia, was part of a seven-member crew aboard the Orbiter Challenger in October 1984 and later served as assistant administrator for congressional relations for NASA. Farkas expressed the hope he one day follow in the footsteps of fellow astronaut John Glenn, who at 77 is about to go into space again. On May 22, 1995, Farkas was made a brigadier general, and the following year he was appointed military attache at the Hungarian embassy in Washington. However, cited by District of Columbia traffic police in December for driving under the influence of alcohol, Farkas was ordered home and retired.    

Question: Is Farkas sent into space on board the Soyuz before McBride on board the Orbiter Challenger?

The step-by-step explanation and answer is: Farkas was sent into space on May 26, 1980 and McBride was on board the Orbiter Challenger in October 1984. May 26, 1980 is before October 1984. So the answer is: yes.

Question: Is McBride on board the Orbiter Challenger after Farkas was made a brigadier general?

The step-by-step explanation and answer is: McBride was on board the Orbiter Challenger in October 1984. Farkas was made a brigadier general on May 22, 1995. October 1984 is before May 22, 1995. So the answer is: no.

Question: Is Farkas was appointed military attache at the Hungarian embassy before he was made a brigadier?

The step-by-step explanation and answer is: Farkas was made a brigadier general on May 22, 1995. He was appointed military attache at the Hungarian embassy in the following year. So the answer is: no.}

{\color{TealBlue} Answer the question according to the article. Give step-by-step explanations and then answer yes or no. 

Article: \{events\}. 

Question: \{question\}. 

The step-by-step explanation and answer is:} \\
\bottomrule
\end{tabular}
\caption{Example model input using Chain-of-thought prompts for TempEvalQA-Bi. The input consists of \textit{\color{Orange} demonstrations} and {\color{TealBlue} the actual prompt}.}
\label{tab:prompt-cot}
\end{table*}

\begin{table*}[t!]
\begin{tabular}{p{0.99\linewidth}}
   \toprule 
  \textbf{Event:} \textit{Well. I started this blog years ago and never kept up with it. In fact, I sort of forgot about it. Now, I am done with school and married and I actually have time on my hands! (Gasp! Imagine that!) And, I really want to get back into the swing of writing, so I thought blogging might be a good way to start. I feel like an introduction is in order, so here's a bit about me. My name is Kathleen. I am 22 years old, and I live in Portland, Maine with my husband Brett. We are originally from Portland, Oregon but lived in Seattle for 4 years during college. I work as an office manager in a small software company, so I tend to have a bit of time on my hands. Hence, the blogging. I'll probably talk a lot about all of the changes that have happened in my life in the past year, since that's what is on my mind right now: getting married in July, moving to Maine in November, changing jobs several times, etc. So I will get right down to it. I am pretty freaking grumpy today. Well, to be honest, I've been really grumpy for a few weeks and I can't seem to get un -grumpy, which I hate. Feeling out of control, especially of myself, is not something I like. I guess I'm coming to realize that life, and especially marriage, is NOT easy. Maybe I assumed that since we dated for over 4 years before we got married, that things would be easy. Everyone says that the first year of marriage is the hardest, and for the first six months I laughed that off. ""Hard?"" I would scoff. ""No way! I love being married. It's wonderful! It's amazing! blah, blah, blah."" Now it's like the crap has hit the fan and I am transformed into a grumpy, nagging, mean, yelling wife. Ugh. I don't want to be like this but every time I resolve to be different I fall right back into it. Choosing your battles is a lot harder than it seems. I think that's a lot of what I'm learning; that I really do have to choose my battles, because if I pick on every little thing, he won't listen at all when the big things come along and I really need him to listen. Heh. That's my psychological epiphany for the day. } \\
  \midrule
  \textbf{Q:} Did the getting married happen after they dated? \\
\\
  \textbf{LLaMA-65B:} \hlblue{They dated for over 4 years before they got married. They dated for over 4 years happened before they got married.} So the answer is: \hlgreen{yes}. \\
  \\
  \textbf{GPT-4:} \hlblue{Kathleen mentioned that they dated for over 4 years before they got married.} So the answer is \hlgreen{yes}. \\
  \midrule
  
  \textbf{Q:} Did the getting married happen before they dated? \\
\\
  \textbf{LLaMA-65B:} \hlblue{Getting married happened in July. Dating happened before getting married. July is before getting married.} So the answer is: \hlgreen{no}. \\
  \\
  \textbf{GPT-4:} \hlblue{They dated for over 4 years before getting married.} So the answer is \hlgreen{no}.  \\
  \bottomrule
\end{tabular}
\caption{Example of chain-of-thought for LLaMA-65B. Chain-of-thought reasoning processes are \hlblue{highlighted}. We also highlight the answers from the models to indicate whether they are \hlgreen{correct} or \hlred{incorrect}. Both LLaMA-65B and GPT-4 consistently provided correct answers for both questions. }
\label{tab:cot-example-blog}
\end{table*}

\begin{table*}[t!]
\begin{tabular}{p{0.95\linewidth}}
   \toprule 
  \textbf{Event:} \textit{P.F. Chang's China Bistro Inc. has provided new details about a security breach discovered in June, saying customer data may have been stolen from 33 restaurants in 16 states, including eight in California. The restaurant chain said Monday that credit card numbers, expiration dates and, in some cases, cardholder names were stolen over eight months. However, the chain has not yet determined if ""any specific cardholder's credit or debit card data was stolen by the intruder,"" according to Chief Executive Rick Federico. P.F. Chang's confirmed the data breach June 13, three days after the U.S. Secret Service alerted the chain that its credit card processing systems may have been hacked. The company said the breach occurred between Oct. 19, 2013, and June 11, 2014. The intrusion was first reported by security blogger Brian Krebs, who said on his website that banks reported data from thousands of customers had been pilfered from locations in Florida, Maryland, New Jersey, Pennsylvania, Nevada and North Carolina, and was being sold online. Federico said P.F. Chang's conducted an internal investigation with the help of forensic data experts, and determined the 33 locations affected and specific time frames the credit card processing system was compromised for each. He also said the company has been processing card data securely since June 11. } \\
  \midrule
  \textbf{Q:} Did the determined happen before the processing? \\
\\
  \textbf{LLaMA-65B:} \hlblue{1) The determined happened between Oct. 19, 2013 and June 11, 2014. 2) The processing happened after June 11, 2014. 3) Oct. 19, 2013 and June 11, 2014 is before June 11, 2014.} So the answer is: \hlgreen{yes}. \\
  \\
  \textbf{GPT-4:} \hlblue{The determination of the affected locations and time frames happened after the processing was compromised.} So the answer is \hlred{no}. \\
  \midrule
  
  \textbf{Q:} Did the determined happen after the processing? \\
    \\
  \textbf{LLaMA-65B:} \hlblue{The company said the breach occurred between Oct. 19, 2013, and June 11, 2014. The determined was determined on June 13.} So the answer is: \hlred{yes}.  \\
  \\
  \textbf{GPT-4:} \hlblue{The company determined the affected locations and specific time frames after processing the card data securely.} So the answer is \hlred{yes}.  \\
  \bottomrule
\end{tabular}
\caption{Example of chain-of-thought for LLaMA-65B. Chain-of-thought reasoning processes are \hlblue{highlighted}. We also highlight the answers from the models to indicate whether they are \hlgreen{correct} or \hlred{incorrect}. LLaMA-65B successfully answered the first question accurately but exhibited inconsistency in answering the second question. On the other hand, GPT-4 failed to provide correct answers for both questions.}
\label{tab:cot-example-news}
\end{table*}

\begin{table*}[t!]
\begin{tabular}{p{0.95\linewidth}}
   \toprule 
  \textbf{Event:} \textit{The Great Depression was a severe worldwide economic depression in the decade preceding World War II. The timing of the Great Depression varied across nations, but in most countries it started in 1930 and lasted until the late 1930s or middle 1940s. It was the longest, deepest, and most widespread depression of the 20th century. In the 21st century, the Great Depression is commonly used as an example of how far the world's economy can decline. The depression originated in the U.S., after the fall in stock prices that began around September 4, 1929, and became worldwide news with the stock market crash of October 29, 1929 (known as Black Tuesday). The Great Depression had devastating effects in countries rich and poor. Personal income, tax revenue, profits and prices dropped, while international trade plunged by more than 50\%. Unemployment in the U.S. rose to 25\%, and in some countries rose as high as 33\%. Cities all around the world were hit hard, especially those dependent on heavy industry. Construction was virtually halted in many countries. Farming and rural areas suffered as crop prices fell by approximately 60\%. Facing plummeting demand with few alternate sources of jobs, areas dependent on primary sector industries such as cash cropping, mining and logging suffered the most. Some economies started to recover by the mid-1930s. In many countries, the negative effects of the Great Depression lasted until after the end of World War II. Start. Economic historians usually attribute the start of the Great Depression to the sudden devastating collapse of US stock market prices on October 29, 1929, known as Black Tuesday; some dispute this conclusion, and see the stock crash as a symptom, rather than a cause, of the Great Depression. \ldots  } \\
  \midrule
  \textbf{Q:} Did the depression start before middle 1940s? \\
  \\
  \textbf{LLaMA-65B:} \hlblue{The depression started after the fall in stock prices that began around September 4, 1929. 2. Middle 1940s is after September 4, 1929.} 3. So the answer is: \hlred{no}. \\
  \\
  \textbf{GPT-4:} \hlblue{The Great Depression started in the decade preceding World War II, and in most countries, it began in 1930. It lasted until the late 1930s or middle 1940s.} So the answer is \hlgreen{yes}. \\
  \midrule
  
  \textbf{Q:} Did the depression start after middle 1940s? \\
  \\

  \textbf{LLaMA-65B:} \hlblue{The depression originated in the U.S., after the fall in stock prices that began around September 4, 1929, and became worldwide news with the stock market crash of October 29, 1929 (known as Black Tuesday).} So the answer is: \hlgreen{no}.  \\
  \\
  \textbf{GPT-4:} \hlblue{The depression started in the decade preceding World War II, which means it began before the 1940s. In most countries, the Great Depression started in 1930 and lasted until the late 1930s or middle 1940s.} So the answer is \hlgreen{no}.  \\
  \bottomrule
\end{tabular}
\caption{Example of chain-of-thought for LLaMA-65B. Chain-of-thought reasoning processes are \hlblue{highlighted}. We also highlight the answers from the models to indicate whether they are \hlgreen{correct} or \hlred{incorrect}. LLaMA-65B demonstrated accuracy in answering the second question while displaying inconsistency in answering the first question. Conversely, GPT-4 provided correct responses for both questions.}
\label{tab:cot-example-wikipedia}
\end{table*}

\end{document}

%% file: macros.tex
\usepackage{amsmath,amsfonts,bm}

\def\eqref#1{equation~\ref{#1}}

\def\1{\bm{1}}

\DeclareMathAlphabet{\mathsfit}{\encodingdefault}{\sfdefault}{m}{sl}
\SetMathAlphabet{\mathsfit}{bold}{\encodingdefault}{\sfdefault}{bx}{n}

\usepackage{todonotes}
\makeatletter
\newcommand*\iftodonotes{\if@todonotes@disabled\expandafter\@secondoftwo\else\expandafter\@firstoftwo\fi}
\makeatother

\definecolor{edolime}{rgb}{0.9,1,0.3}